\def\year{2022}\relax
\documentclass[letterpaper]{article} 
\usepackage{aaai22}  
\usepackage{times}  
\usepackage{helvet}  
\usepackage{courier}  
\usepackage[hyphens]{url}  
\usepackage{graphicx} 
\usepackage{natbib}  
\usepackage{caption} 
\DeclareCaptionStyle{ruled}{labelfont=normalfont,labelsep=colon,strut=off} 
\usepackage{algorithm}
\usepackage{algorithmic}

\usepackage{amsmath}
\usepackage{pifont}
\usepackage{booktabs}
\usepackage{newfloat}
\usepackage{listings}
\floatstyle{ruled}
\newfloat{listing}{tb}{lst}{}
\floatname{listing}{Listing}
\title{Dynamic Relation Discovery and Utilization \\ in Multi-Entity Time Series Forecasting}
\author{
    Lin Huang, Lijun Wu, Jia Zhang, Jiang Bian, Tie-Yan Liu
}
\affiliations{
    \textsuperscript{\rm 1}Microsoft Research Asia, BeiJing, China\\

    huang.lin@microsoft.com, lijun.wu@microsoft.com, jia.zhang@microsoft.com, \\jiang.bian@microsoft.com, tie-yan.liu@microsoft.com
}

\usepackage{bibentry}

\begin{document}

\maketitle

\begin{abstract}
Time series forecasting plays a key role in a variety of domains. In a lot of real-world scenarios, there exist multiple forecasting entities (e.g. power station in the solar system, stations in the traffic system). A straightforward forecasting solution is to mine the temporal dependency for each individual entity through 1d-CNN, RNN, transformer, etc. This approach overlooks the relations between these entities and, in consequence, loses the opportunity to improve performance using spatial-temporal relation. However, in many real-world scenarios, beside explicit relation, there could exist crucial yet implicit relation between entities. How to discover the useful implicit relation between entities and effectively utilize the relations for each entity under various circumstances is crucial.
In order to mine the implicit relation between entities as much as possible and dynamically utilize the relation to improve the forecasting performance, we propose an attentional multi-graph neural network with automatic graph learning (A2GNN) in this work. 
Particularly, a Gumbel-softmax based auto graph learner is designed to automatically capture the implicit relation among forecasting entities. We further propose an attentional relation learner that enables every entity to dynamically pay attention to its preferred relations.
Extensive experiments are conducted on five real-world datasets from three different domains. The results demonstrate the effectiveness of A2GNN beyond several state-of-the-art methods.

\end{abstract}

\section{INTRODUCTION}

Time series forecasting is playing a vital role in many application scenarios of a variety of domains, such as solar power generation forecasting in renewable energy \citep{10,11,mtgnn}, traffic forecasting in transportation system \citep{traffic1,traffic2},
electricity consumption forecasting in social life \citep{10,11,mtgnn}, market trend prediction in financial investment~\citep{finance1,finance2}, etc.
In many of them, there exist multiple forecasting entities, e.g. power stations in the solar system, stations in the traffic system, stocks in financial market, and commodities in retailing business.

A straightforward forecasting solution is to mine temporal dependency for each individual entity by 1d-convolution neural network (1D-CNN) \citep{tcn-2018}, recurrent neural network (RNN) \citep{lstm,gru1}, transformer \citep{attention-is-all-you-need}, and etc., while it will overlook important relations between these entities, such as explicit relation defined by human knowledge (e.g. competitive, cooperative, causal, and geospatial relation) and other implicit relation behind the data.
In reality, such relations could provide valuable signals towards accurate forecasting for each individual entity.
For example, in traffic system, the station can affect its geospatial adjacent stations;
the stock price of upstream companies in the supply-chain can substantially indicate that of downstream ones in the financial market scenario; the similarity relation between entities behind the data can also increase the robustness of model. 
Spatial-temporal graph neural networks \cite{dcrnn,stmetanet,graphwavenet,mtgnn}, in which the entity relations are fully utilized, are good examples with better forecasting performance.

In many real-world scenarios, however, beside some explicit relations, there usually exist crucial yet implicit relations between entities.
Recently, a growing number of research works pay attention to graph neural network (GNN) \citep{gcn,graphsage,gat,protein1,physical-system1} to leverage the implicit relations between entities~\citep{ddgf,graphwavenet,Gumbel-graph-neural-network-2019,mtgnn}. 
One common idea among these works is to assume all entities composing a complete graph and let GNNs automatically learn the pairwise correlations between any two entities~\citep{ddgf,graphwavenet}. 
To avoid the computation complexity and over-smoothing issues of above-mentioned studies, the operation analogous to sparse encoding~\citep{mtgnn} has been further proposed.
Nevertheless, an imprudent employment of sparse constraints at the earlier learning phase may limit the model capacity of discovering crucial relation, while the uni-directional relation is not suitable for every real scenario.

Moreover, there also exist critical challenges regarding the utilization of multiple relations, one of which is the information aggregation through different types of relations, especially the co-existence of explicit relation as well as implicit relation.
A straightforward method is to employ different graph convolution blocks for different relation types and then conduct a direct fusion \citep{graphwavenet}. Some others conduct fusion based on correlation between nodes by leveraging the meta information of both nodes and edges \citep{stmetanet}. In addition, Graph multi-attention network (GMAN) \citep{gman} employs a gated fusion mechanism to fuse the spatial and temporal representations. 
However, all of these efforts rely on static fusion of multiple relations reflecting either pre-existed or latent connections. Indeed, given the distinct characteristics of multiple relations, accurate forecasting usually relies on dynamic reliance on them under various circumstances.

\begin{figure}[!htbp]
\centerline{\includegraphics[width=1\linewidth]{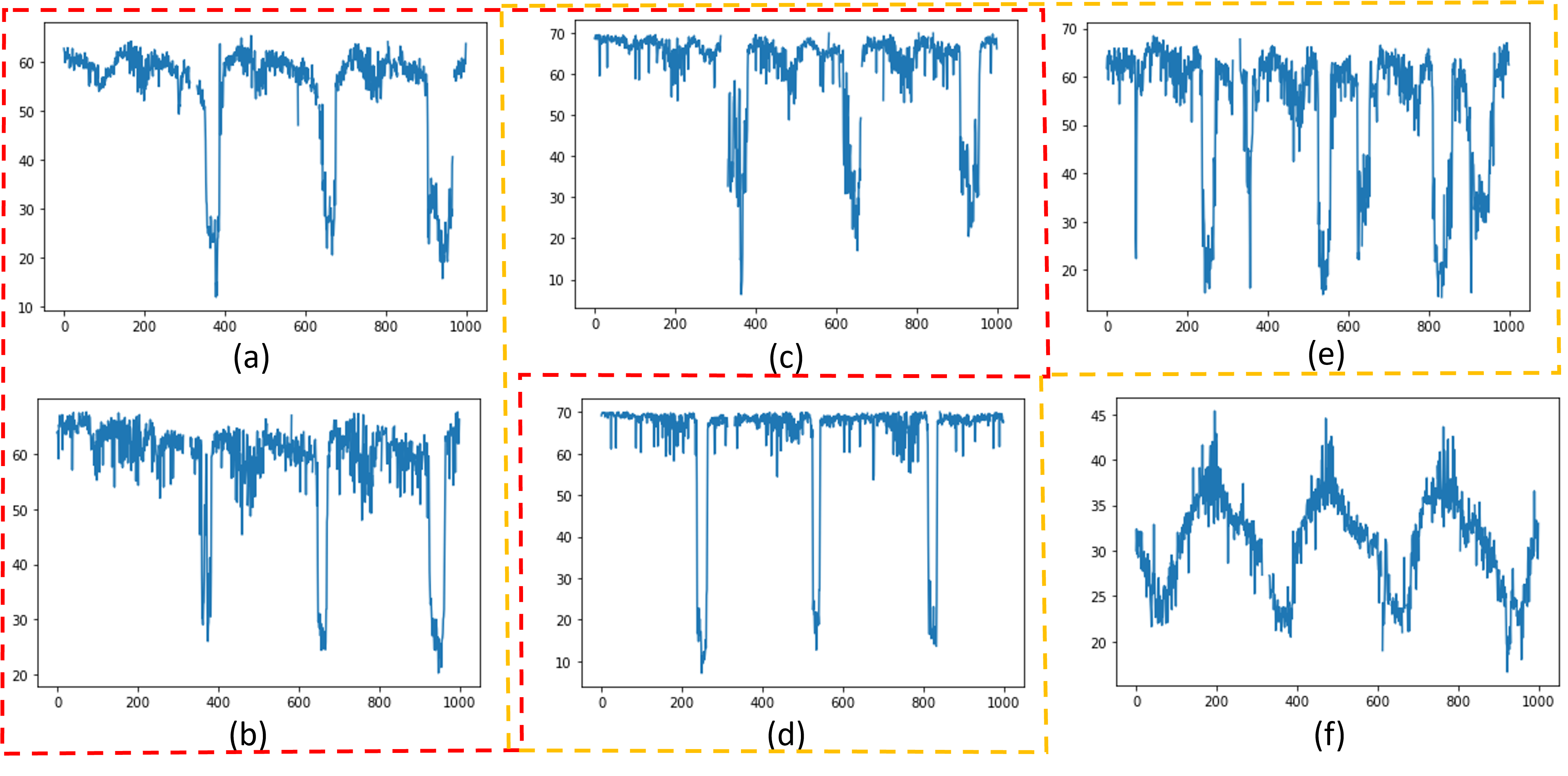}}
\caption{Diagrammatic representation of the relationships between entities.}
\label{fig:Diagrammatic representation}
\end{figure}

Thus, the problem lies in two crucial aspects, i.e., implicit relation {\em discover} and relations' effective {\em utilization}.
We take the $6$ entities in Fig. \ref{fig:Diagrammatic representation} to illustrate the importance of them.
In Fig. \ref{fig:Diagrammatic representation}, entity-(a), entity-(b), and entity-(c) have a similar curve, thus, the similarity can be employed to enhance the robustness of the model.
Although there does not exist a similar curve for entity-(e), it is not hard to see that when entity-(c) or entity-(d) is in lowest peak, entity-(e) is also in lowest peak. Thus, entity-(e) can use this relation for better forecasting.
The periodicity of entity-(f) is strong so that only employing its own information can also get an accurate forecasting result (i.e. the explicit and implicit relation may be not necessary for entity-(f)).

In this paper, in order to mine the implicit relation between entities as much as possible and effectively utilize the relations to improve the forecasting performance, we propose an attentional multi-graph neural network with automatic graph learning (A2GNN).
In particular, we propose a novel auto graph learner based on Gumbel-Softmax~\citep{gumbelsoftmax} to sample all feasible entity pairs so that the relation between any entity pair has the chance to be reserved in the learned graph, while we leverage the sparse matrix to ensure the computing efficiency. Moreover, we propose an attentional relation learner so that every entity can dynamically pay attention to useful relations, resulting in more flexible utilization of multiple relations and consequently better forecasting performance.



The main contributions of this paper include:
\begin{itemize}
\item We propose a new graph neural network framework, A2GNN, with automatic discovery and dynamic utilization of relations for time series forecasting.
\item Within A2GNN, a novel auto graph learner based on Gumbel-Softmax~\citep{gumbelsoftmax} can effectively discover the implicit relation between entities while significantly reducing complexity.
\item Within A2GNN, an attentional relation learner enables every entity dynamically pay more attention to their preferable relations, resulting in more flexible utilization of multiple relations.
\item 
Our proposed A2GNN is quite general such that it can be applied to both time series and spatial-temporal forecasting tasks.
And, extensive experiments have shown that our method outperforms the state-of-the-art methods on a couple of well-known benchmark datasets.
\end{itemize}

\section{RELATED WORK}
\label{related work}
Our work is related to three lines of research: time series forecasting methods, spatial-temporal graph neural networks, and graph neural networks sparsification methods.

\subsection{Time Series Forecasting Methods}

There are plenty of works on time series forecasting problem \citep{5,7,8,gru1,gru2,lstm}. Recently, long and short-term time-series network (LSTNet) \citep{10} utilizes the convolution neural network (CNN) \citep{cnn} and recurrent neural network (RNN) to extract short-term local dependency and long-term patterns for time series trends. Shih proposes a temporal pattern attention (TPA-LSTM)\citep{11} to select relevant time series, and leverages its frequency domain information for multivariate forecasting. A Lot of studies \citep{unfoldingtemporaldynamics-2016-aaai,restful-wu-2018,dsanet-huang-2019} also focus on multi-scale temporal information extraction. Existing methods focus more on time series information utilization for forecasting, but they neglect the implicit relation among entities. 


\subsection{Spatial-Temporal Graph Neural Networks}

A spatial-temporal forecasting task has pre-defined relation (i.e. explicit relation) by human knowledge, and traffic forecasting is a typical and hot problem of spatial-temporal forecasting as the natural geographic relation. Similar like time series task, a lot of works focus on temporal information mining, such as deep spatio-temporal residual networks (ST-ResNet) \citep{stresnet}, spatio-temporal graph convolutional network (STGCN) \citep{stgcn}, Graph-WaveNet \citep{graphwavenet,wavenet}. Some works focus on natural geographic relation utilization, such as diffusion convolutional recurrent neural network (DCRNN) \citep{dcrnn}, spatial-temporal forecasting with meta knowledge (ST-MetaNet) \citep{stmetanet}, graph multi-attention network (GMAN) \citep{gman}, multi-range attentive bicomponent GCN (MRA-BGCN) \citep{mrabgcn}, spatial temporal graph neural network (STGNN) \citep{STGNN},  Spectral temporal graph neural network (StemGNN) \citep{StemGNN}, and etc. 



Implicit relation discover aroused researcher's attention. Graph convolutional neural network with data driven graph filter (DDGF) \citep{ddgf} breaks this limitation and discovers implicit relation to replace the pre-defined relation, by calculating all pairwise correlations between nodes. However, the risk of over smoothing and computation complexity increase when taking all nodes as neighbors. 
Gumbel Graph Network (GGN) \citep{Gumbel-graph-neural-network-2019}  proposes a model-free, data-driven deep learning framework to accomplish the reconstruction of network connections, while the one time sampler will curb the efficiency of connection reconstruction.
Multivariate time graph neural network (MTGNN)~\citep{mtgnn} accelerates the computation efficiency in traffic forecasting by employing a sparse uni-directional graph to learn hidden spatial dependencies among variables. Nevertheless, a imprudent employment of sparse constraints at the earlier learning phase may limit the model capacity of discovering crucial relation and the assumption of uni-direction limits its application in many real scenarios. 


\subsection{Graph Neural Networks Sparsification Methods}
Graph sparsification aims at finding small subgraphs from given implicit large graphs that
best preserve desired properties. 
For instance, Fast learning with graph neural networks (FastGCN) \citep{fastgcn-2018-ICLR}  interpret graph convolutions as integral transforms of embedding functions under probability measures and uses Monte Carlo approaches to consistently estimate the integrals.
NeuralSparse \citep{graphsparse-2020-ICML} considers node/edge features as parts of input and optimizes graph sparsification by supervision signals from errors made in downstream tasks.

\begin{figure*}[ht!]
\centerline{
\includegraphics[width=0.75\linewidth]{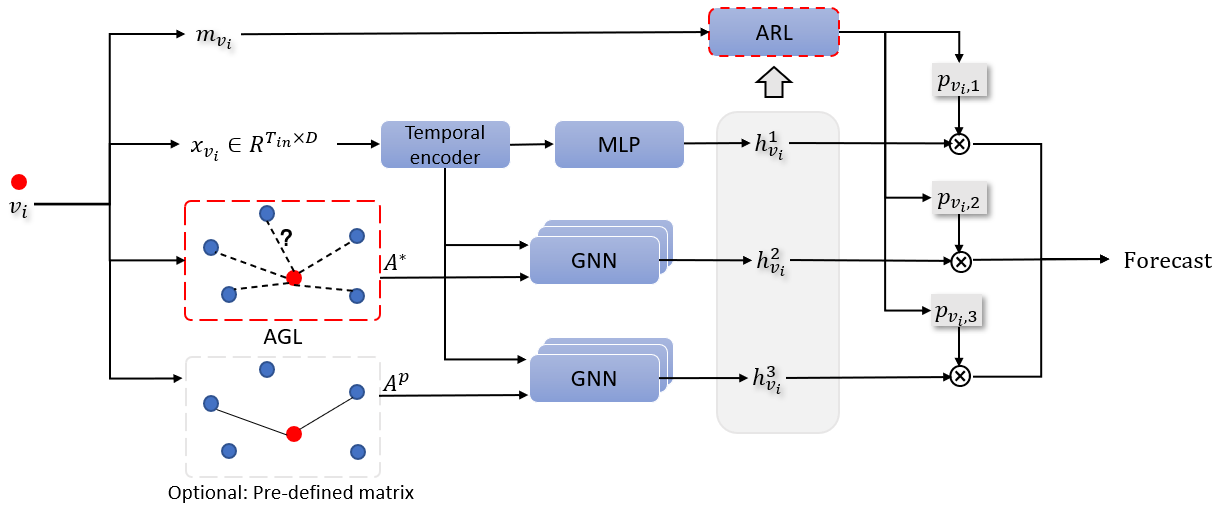}
}
\caption{The framework of A2GNN. AGL is the auto graph learner for implicit graph discover. ARL is the attentional relation learner that enables each entity dynamically pay attention to different relations. $\mathbf{m}_{v_i}$ is the learnable node embedding of node $v_i$. $\mathbf{x}_{v_i}$ is historical observed features. 
$\mathbf{h}^1_{v_i}$, $\mathbf{h}^2_{v_i}$, and $\mathbf{h}^3_{v_i}$ mean own temporal information, neighbor information by implicit relation, and neighbor information by explicit relation, respectively. $p_{v_i,1}$, $p_{v_i,2}$, and $p_{v_i,3}$ are  dynamic weight given by ARL.}
\label{fig:framework of A2GNN}
\end{figure*}

\section{PROBLEM FORMULATION}
In this section, we first provide a detailed problem formulation. Suppose there are $N$ forecasting entities, we use $x_t\in R^{N\times d}$ to stand for the values of these $N$ entities at time step $t$, and $d$ is the feature dimension. $y_{t}\in R^{N\times 1}$ is the feature/variable need to forecast. The historical observations of past $t_{in}$ steps before $t$ is defined as $X=\{x_{t-t_{in}+1},x_{t-t_{in}+2},\cdots,x_t \}$.
Our goal is to build a function to predict a sequence of values $Y=\{y_{t+1},y_{t+2},... ,y_{t+t_{out}}\}$ for future $t_{out}$ steps.

We give formal definitions of graph-related concepts:

\textbf{\textit{Graph: }}
A graph is formulated as $G=(V,E)$ where $V$ is the set of nodes, and $E$ is the set of edges. We use $N=|V|$ to denote the number of nodes in the graph. In particular, in forecasting scenario, we consider each forecasting entity as a node in the graph.

\textbf{\textit{Node neighbors: }}
For any node $v_i,v_j\in V$, $(v_i,v_j)\in E$ denote an edge between $v_i$ and $v_j$. The neighbors of $v_i$ are defined as $N_{v_i}=\{v_k ~|~ (v_k,v_i)\in E \land v_k\in V \}$.

\textbf{\textit{Weighted adjacency matrix: }}
The weighted adjacency matrix is a mathematical representation of the relation. We denote the weighted adjacency matrix as $A\in R^{N\times N}$, where $A_{i,j}>0$ if there is an edge between the node $v_i$ and node $v_j$, otherwise, $A_{ij}=0$. The weight $A_{i,j}$ stands for some metrics of relation between these two nodes. 

\section{Proposed Framework}
In this section, we introduce the proposed attentional multi-graph neural network with automatic graph learning (A2GNN). 
As shown in Figure~\ref{fig:framework of A2GNN}, the whole framework of A2GNN consists of 5 parts: temporal encoder, auto graph learner (AGL), graph neural network (GNN), and attentional relation learner (ARL). The key parts of this study are AGL and ARL. In the end, we employ the multi-layer perception (MLP) for efficient inference instead of RNN.

The algorithm is shown in Alg. \ref{alg:algorithm}. 
It can be easily applied both spatial-temporal and time series forecasting tasks, and the difference between them is whether exist pre-defined relation. For better understand A2GNN, we make a brief introduction about its temporal encoder and a detailed introduction about the key parts: AGL and ARL.

\newcommand{\parag}[1]{\vspace{1mm}\textit{#1}:\ }
\parag{Temporal encoder} 1D-CNN~\citep{tcn-2018,cnn}, RNN~\citep{gru1,gru2,lstm}, and transformer~\citep{attention-is-all-you-need}, can be employed to extract the temporal information for each node.
The representation will be used as input for the further graph neural network. 
In this paper, we use LSTM to extract temporal information.

\begin{algorithm}[ht]
\caption{A2GNN}
\label{alg:algorithm}
\textbf{Input}: Node temporal information $X\in R^{N\times T_{in} \times D}$; Pre-defined graph $A^p \in R^{N\times N}$(Optional).\\
\textbf{Parameter}: Node embedding $M\in R ^ {N \times D_m}$; Random initialized adjacent matrix $A\in R^{N \times N}$; Parameter of graph neural network; Parameter of auto relation learner.\\
\textbf{Output}: $\hat{Y}\in R^{N\times T_{out} \times D}$
\begin{algorithmic}[1] 
\STATE{/* Discover implicit relation */}
\IF {training} 
\STATE $A^* = \textsc{AGL}_{training}(A)$
\ELSIF{inference}
\STATE $A^* = \textsc{AGL}_{inference}(A)$
\ENDIF
\STATE Extract temporal information: $S=\textsc{LSTM}(X)$
\STATE Extract node own information: $H^1=\textsc{MLP}(S)$
\STATE Aggregate implicit neighbor: $H^2=\textsc{GNN}(S,A^*)$
\STATE Aggregate pre-defined neighbor: $H^3=\textsc{GNN}(S,A^p)$
\STATE Utilize multiple relations: $Z=\textsc{ARL}(M,H^1,H^2,H^3)$
\STATE /* efficient inference */
\STATE $Y=ZW$
\STATE \textbf{return} $\hat{Y}$
\end{algorithmic}
\end{algorithm}

\subsection{Auto Graph Learner} 
\begin{figure}[t]
    \centering
    \includegraphics[width=1\linewidth]{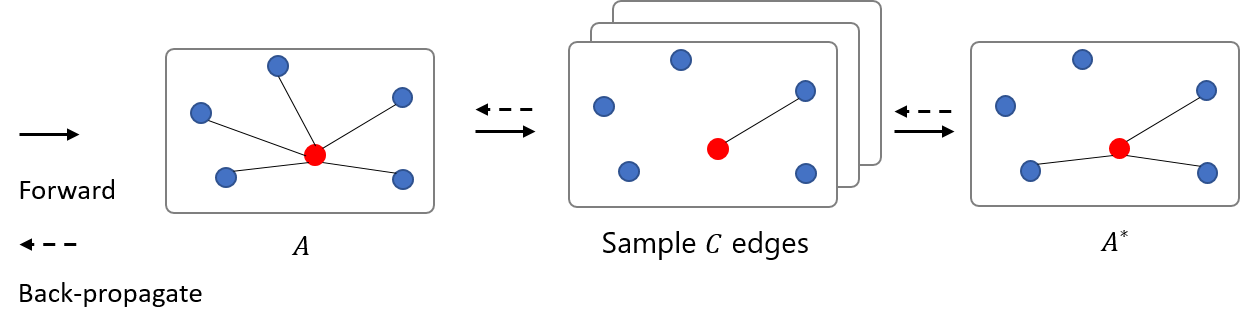}
    \caption{The process of auto graph learner in training. }
    \label{fig:auto graph learner process}
\end{figure}

Auto graph learner (AGL) kicks off the learning for implicit relation from a randomly initiated adjacency matrix. Towards minimizing the over smoothing risk, we use sample-based method to select potential edges. Moreover, multiple edges of each node are sampled for an efficient training.
However, the vanilla sample operation is not differentiable, which hinders the back propagation for edges' weight optimization. To address this issue, we employ Gumbel-Softmax \citep{gumbelsoftmax} method to make a differentiable sampling. There are some differences in training and inference phases and the detailed procedures are shown in below.

\parag{Training} Given the randomly initiated adjacency matrix $A\in R^{N\times N}$, the calculation steps include: 

1. $\forall v_j\in N_{v_i}$, we employ a softmax function to compute the sampling probability of each edge:
\begin{equation}
    \pi_{v_i,v_j}=\frac{\exp(a_{v_i,v_j})} {\sum_{v_k\in N_{v_i}} \exp(a_{v_i,v_k})},
    \label{eqn:matrix softmax}
\end{equation}
where $\pi_{v_i,v_j}$ is the weight of edge $(v_i,v_j)$ after softmax, and $a_{v_i,v_j}$ is the weight of edge $(v_i,v_j)$ in adjacency matrix $A$. 

2. We then generate differentiable samples through Gumbel-Softmax:
\begin{equation}
			\pi_{v_i,v_j}^{G}=\frac{\exp( (\log(\pi_{v_i,v_j})+\epsilon_{v_j})/\tau )}   {\sum_{v_k\in N_{v_i}} \exp( (\log(\pi_{v_i,v_k})+\epsilon_{v_k})/\tau )},
\label{eqn:gumbel softmax}
\end{equation}
where $\pi_{v_i,v_j}^{G}$ is the weight of edge $(v_i,v_j)$ after Gumbel-Softmax.  $\epsilon=-\log(-\log(s))$, with randomly generated $s$ from Uniform distribution U$(0,1)$, and $\tau \in {(0,+\infty)}$ is a hyper-parameter called temperature. As the softmax temperature $\tau$ approaches $0$, samples from the Gumbel-Softmax distribution approximate one-hot vector, which means discrete. 

3. Repeating above procedure $C$ times, we can obtain $C$ samples. Finally, we apply normalization over all samples to get the final correlations for all nodes:
\begin{equation}
			a^*_{v_i,v_j}= \frac{\sum_{c=1}^{C} \pi ^{G,c}_{v_i,v_j} }   {\sum_{c=1}^{C}\sum_{v_k\in N_{v_i}} \pi^{G,c}_{v_i,v_k}},
\label{eqn:training rate}
\end{equation}
where $a^*_{v_i,v_j}$ is the weight of edge $(v_i,v_j) $and will be further used in the calculation procedures in the graph neural network. $\pi ^{G,c}$ is the $c$-th Gumbel-Softmax result.

\parag{Inference} Since the auto graph learner has already learned the nodes' relation by updating matrix $A$ in the training phase, we replace Gumbel-Softmax sample operation by selecting the top related $C$ neighbors $\textsc{top}_C (N_{v_i})$ for each node $v_i$ (the edge weight is used as the metric for $\textsc{top}$ operation). Then softmax is applied to normalize these edge weights:
\begin{equation}
    a^*_{v_i,v_j}=\frac{\exp(a_{v_i,v_j})} {\sum_{v_k\in \textsc{top}_C (N_{v_i})} \exp(a_{v_i,v_k})}, ~~v_j\in \textsc{top}_C (N_{v_i})
    \label{eqn:evaluation matrix softmax}
\end{equation}

The $C$ times sampling operation ensures the rationality and efficacy of relation learning process.
At the begin of training, all the edges have the probability to be sampled, and the sampled edge's weight can be properly updated.
As time goes on, the useful edges will have larger weights through the updating with back propagation algorithm.

The output of AGL is $A^*\in R^{N\times N}$, and the element of which is $a^*_{v_i,v_j}$.

\subsection{Graph Neural Network}
The node's time series information is processed by LSTM before input into the graph neural network. Specifically, the hidden state from LSTM in each time step will be concatenated as the output. The output from LSTM block (i.e. temporal encoder) is defined as $S=\textsc{LSTM}(X)$, and $S\in R^{N\times d_{enc}}$. 



With the assistance of the output from LSTM, the node itself information is processed by multi-layer perceptron (MLP) like $H^1=\textsc{MLP}(S)$.
The neighbor information by implicit relation is aggregated by graph neural network like 
$H^2=A^* \cdots (A^* S W^{(1)}_2) \cdots W^{(l)}_2$, and $l$ is the layer depth, $W$ is learnable parameter for feature transportation. For some scenarios, such as traffic, there always exists natural geographic relation. Thus, the pre-defined natural relation can also be employ by graph neural network like $H^3=A^p \cdots (A^p S W^{(1)}_3) \cdots W^{(l)}_3$, and $A^p$ us the pre-dedined weighted adjacency matrix.

Thus, after the graph neural network, for each node $v_i$, the own information $h^1_{v_i}$, the aggregated information based on implicit relation $h^2_{v_i}$, and the aggregated information based on explicit relation $h^3_{v_i}$ are available from $H^1$, $H^2$, and $H^3$, respectively.

\subsection{Attentional Relation Learner}

The representations $H^1$, $H^2$, and $H^3$ are fed into the attentional relation learner (ARL).  Unlike previous studies that always rely on static fusion of multiple relations (e.g. explicit or implicit relation), AGL enables each node dynamically rely on different relations under various circumstances. Furthermore, node itself information, which can be called own relation, is maintained through $H^1$ instead of skip connection.

\begin{figure}[t]
    \centering
    \includegraphics[width=1\linewidth]{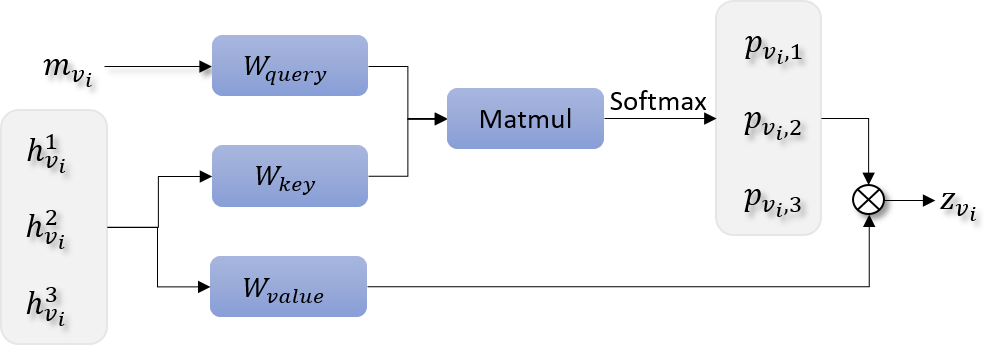}
    \caption{The process of attentional relation learner.}
    \label{fig:attentional relation learner process}
\end{figure}

As show in Fig. \ref{fig:attentional relation learner process}, in ARL, each node has a random initialized embedding vector, that is $M\in R^{N\times d_m}$. 
Each node learns its attention coefficient on multiple representations, that is $H^1$, $H^2$, and $H^3$. In particular, in Eq. \eqref{eqn:attentional relation coefficient}, the dot production and $\textsc{Softmax}$ operation are employed to calculate the attention coefficient.

\begin{equation}
    p_{v_i,k\in \{1,2,3\}} =\textsc{Softmax}_k\left( \frac{(m_{v_i}W_{query})\cdot(h^k_{v_i}W_{key}) }{\sqrt{d}}\right)
    \label{eqn:attentional relation coefficient}
\end{equation}
where $p_{v_i,k\in \{1,2,3\}}$ is the attention coefficient scalar of node $v_i$ on the representation $h^k_{v_i}$. $m_{v_i}$ is node $v_i$'s embedding from $M$. $h^k_{v_i}$ is node $v_i$'s representation from corresponding $H^k$. $W_{query} \in R^{d_m \times d}$ and $W_k \in R^{d_h \times d}$ are learnable parameters for feature transportation. $d_m$ is the dimension of $m_{v_i}$, $d_h$ is the dimension of $h^k_{v_i}$, $d$ is the output dimension of $W_{query}$ and $W_{key}$. $\cdot$ means dot production.



After that, the attention coefficient scalar is employed to merge multiple representations for each node $v_i$ as:
\begin{equation}
			z_{v_i} = \textsc{Concat}_{k\in \{1,2,3\}}\left(p_{v_i}(h^k_{v_i}W_{value}) \right)
\label{eqn: attentional relation learner attention concat}
\end{equation}
where $W_{value}$ is the learnable parameter.

\subsection{Optimizaiton}
For each node $v_i$, final forecasting module will use the $z_{v_i}$ to predict, that is $\hat{y}_{v_i} = z_{v_i}W$. The Root Mean Squared Error (RMSE) is employed as the loss function.

\section{Experiments}
We evaluate the proposed A2GNN framework in $5$ datasets: Solar-energy, Traffic, Electricity, METR-LA, and PEMS-BAY.
Particularly, the first $3$ datasets are used to prove that A2GNN can be employed to model time series forecasting problem, and they have no pre-defined relation. And then, we make further experiments on $2$ well known spatial-temporal traffic datasets compared with state-of-the-art graph neural network method, and these datasets have pre-defined relation (i.e. natural geographic relation).  Details of these datasets are introduced in appendix.

To  evaluate  the  model  performances,  we adopt five metrics, which are Mean Absolute Error (MAE), Root Mean  Squared  Error  (RMSE),  Mean  Absolute  Percentage  Error(MAPE), Relative Squared Error (RSE) and Empirical Correlation Coefficient (CORR). The mathematical formulas are shown in appendix.
All the experiments $5$ times and report the average score in order to remove the influence of randomness (e.g. instability of Gumbel-Softmax operation and randomness of model initialized parameters). More settings of our experiments are shown in appendix.

\subsection{Baseline Methods for Comparison}
As we mentioned above, the biggest difference between time series forecasting task and spatial-temporal forecasting task lies in whether there exists a pre-defined relation. 
All the methods mentioned in Section Related Work are concluded. The details of these baselines are shown in the appendix.



\subsection{Result Comparison}
\subsubsection{Result Comparison on Time Series Dataset}
We compare the performances of the proposed A2GNN model with above-mentioned baseline methods on $3$  well-known time series forecasting datasets, and we want to prove that our method can find the implicit relation and make a better prediction compared with other previous time series forecasting methods.

\begin{table}[ht!]
\renewcommand\arraystretch{1.05}         
\renewcommand\tabcolsep{2.0pt}

\caption{Experiments on time series forecasting datasets.}
\begin{center}
\scalebox{0.66}{
\begin{tabular}{cc|ccc|ccc|ccc}
\toprule
\multicolumn{2}{c|}{Dataset} & \multicolumn{3}{c|}{Solar-Energy} & \multicolumn{3}{c|}{Traffic} & \multicolumn{3}{c}{Electricity} \\ \hline

\multicolumn{2}{c|}{ } & \multicolumn{3}{c|}{ Horizon} & \multicolumn{3}{c|}{ Horizon} & \multicolumn{3}{c}{ Horizon} \\ \hline

\hline
Methods & Metrics & 6 & 12 & 24  & 6 & 12 & 24 & 6 & 12 & 24\\ \hline
AR & RSE$\downarrow$  & 0.379 & 0.591 & 0.869  & 0.621 & 0.625 & 0.630  & 0.103 & 0.105 & 0.105\\
 & CORR$\uparrow$  & 0.926 & 0.810 & 0.531  & 0.756 & 0.754 & 0.751  & 0.863 & 0.859 & 0.859\\ \hline
VAR-MLP & RSE$\downarrow$  & 0.267 & 0.424 & 0.684  & 0.657 & 0.602 & 0.614  & 0.162 & 0.155 & 0.127\\
 & CORR$\uparrow$  & 0.965 & 0.905 & 0.714  & 0.769 & 0.792 & 0.789 & 0.838 & 0.819 & 0.867\\ \hline
GP & RSE$\downarrow$  & 0.328 & 0.520 & 0.797  & 0.677 & 0.640 & 0.599  & 0.190 & 0.162 & 0.127\\
 & CORR$\uparrow$  & 0.944 & 0.851 & 0.597 & 0.740 & 0.767 & 0.790  & 0.833 & 0.839 & 0.881\\ \hline
RNN-GRU & RSE$\downarrow$  & 0.262 & 0.416 & 0.485  & 0.552 & 0.556 & 0.563  & 0.114 & 0.118 & 0.129\\
 & CORR$\uparrow$  & 0.967 & 0.915 & 0.882  & 0.840 & 0.834 & 0.830 & 0.862 & 0.847 & 0.865\\ \hline
LSTNet & RSE$\downarrow$ & 0.255 & 0.325 & 0.464  & 0.489 & 0.495 & 0.497 & 0.093 & 0.100 & 0.100\\
 & CORR$\uparrow$ & 0.969 & 0.946 & 0.887 & 0.869 & 0.861 & 0.858 & 0.913 & 0.907 & 0.911\\ \hline
TPA-LSTM  & RSE$\downarrow$ & 0.234 & 0.323 & 0.438 &  0.465 & 0.464 & 0.476 & 0.091 & 0.096 & 0.100\\
& CORR$\uparrow$ &  0.974 & 0.948 & 0.908 & 0.871 & 0.871 & 0.862 & 0.933 & 0.925 & 0.913\\ \hline
MTGNN & RSE$\downarrow$ & 0.234 & 0.310 & 0.427 & 0.475 & 0.446 & 0.453 & 0.087 & 0.091 & \textbf{0.095}\\
 & CORR$\uparrow$ &  0.972 & 0.950 & 0.903 & 0.866 & 0.879 & 0.881  & 0.931 & 0.927 & 0.923\\ \hline \hline
 A2GNN & RSE$\downarrow$  &\textbf{ 0.223} &\textbf{ 0.288} & \textbf{0.407} & \textbf{0.427}  &\textbf{ 0.437 }& \textbf{0.448} &  \textbf{0.0858} &\textbf{ 0.0903} & 0.0970\\ 
(ours) & CORR$\uparrow$ &\textbf{ 0.976} &\textbf{ 0.958} &\textbf{ 0.910}  &\textbf{ 0.890 }& \textbf{0.885} & \textbf{ 0.881}  & \textbf{0.934 }& \textbf{0.929} & \textbf{0.930}\\
\bottomrule
\multicolumn{11}{l}{$\downarrow$ means lower is better and $\uparrow$ means higher is better} \\
\end{tabular}}
\vspace{-0.5cm}
\label{tab:one step forecasting}
\end{center}
\end{table}

From the Table \ref{tab:one step forecasting}, we can see that A2GNN achieves best performance over almost all time steps on Solar-Energy, Traffic, and Electricity data. In particular, compared to previously state-of-the-art methods, A2GNN can achieve significant improvements in terms of the RSE score on Solar-Energy dataset with $4.6\%$, $7.3\%$ and $4.5\%$ error reduction when $t_{out}$ is set to $6$, $12$ and $24$, respectively. Furthermore, on Traffic dataset, the error reduction in terms of RSE is $10.1\%$, $1.8\%$ and $1.0\%$ correspondingly. 

\begin{table}[ht]
\renewcommand\arraystretch{1.05}         
\renewcommand\tabcolsep{2.0pt}
\caption{Experiments on spatial-temporal forecasting datasets.}
\begin{center}
\scalebox{0.70}{

\begin{tabular}{c|ccc|ccc|ccc}
\toprule
\multicolumn{1}{c|}{Dataset} & \multicolumn{9}{c}{METR-LA} \\  \hline

\multicolumn{1}{c}{} & \multicolumn{3}{|c|}{Horizon 3 } & \multicolumn{3}{|c|}{Horizon 6} & \multicolumn{3}{|c}{Horizon 12}\\

\hline
Methods/Metrics & MAE & RMSE & MAPE & MAE & RMSE & MAPE & MAE & RMSE & MAPE \\ \hline
DCRNN & 2.77 & 5.38 & 7.30\% & 3.15 & 6.45 & 8.80\% & 3.60 & 7.60 & 10.50\%   \\
STGCN & 2.88 & 5.74 & 7.62\% & 3.47 & 7.24 & 9.57\% & 4.59 & 9.40 & 12.70\% \\
Graph-WaveNet & 2.69 & 5.15 & 6.90\% & 3.07 & 6.22 & 8.37\% & 3.53 & 7.37 & 10.01\%\\
ST-MetaNet & 2.69 & 5.17 & 6.91\% & 3.10 & 6.28 & 8.57\% & 3.59 & 7.52 & 10.63\%\\
MRA-BGCN & 2.67 & 5.12 & 6.80\% & 3.06 & 6.17 & 8.30\% & 3.49 & 7.30 & 10.00\%\\
GMAN & 2.77 & 5.48 & 7.25\% & 3.07 & 6.34 & 8.35\% & 3.40 & 7.21 & 9.72\%\\
MTGNN & 2.69 & 5.18 & 6.86\% & 3.05 & 6.17 & 8.19\% & 3.49 & 7.23 & 9.87\%\\
A2GNN (ours)& \textbf{2.63} & \textbf{4.99} & \textbf{6.74\%} & \textbf{2.95} & \textbf{5.95} &\textbf{ 8.02}\% & \textbf{3.34} & \textbf{7.00} & \textbf{9.65}\% \\ \hline
Dataset & \multicolumn{9}{c}{PEMS-BAY} \\ \hline
DCRNN& 1.38 & 2.95 & 2.90\% & 1.74 & 3.97 & 3.90\% & 2.07 & 4.74 & 4.90\% \\
STGCN& 1.36 & 2.96 & 2.90\% & 1.81 & 4.27 & 4.17\% & 2.49 & 5.69 & 5.79\% \\
Graph-WaveNet& 1.30 & 2.74 & 2.73\% & 1.63 & 3.70 & 3.67\% & 1.95 & 4.52 & 4.63\% \\
ST-MetaNet& 1.36 & 2.90 & 2.82\% & 1.76 & 4.02 & 4.00\% & 2.20 & 5.06 & 5.45\% \\
MRA-BGCN& 1.29 & 2.72 & 2.90\% & 1.61 & 3.67 & 3.80\% & 1.91 & 4.46 & 4.60\% \\
GMAN& 1.34 & 2.82 & 2.81\% & 1.62 & 3.72 & 3.63\% & 1.86 & 4.32 &\textbf{ 4.31}\% \\
MTGNN& 1.32 & 2.79 & 2.77\% & 1.65 & 3.74 & 3.69\% & 1.94 & 4.49 & 4.53\% \\
A2GNN (ours)& \textbf{1.28} & \textbf{2.70} & \textbf{2.72}\% &\textbf{ 1.58} & \textbf{3.62} & \textbf{3.61}\% & \textbf{1.85} & \textbf{4.29} & 4.39\%\\
\bottomrule  

\multicolumn{10}{c}{%
  \begin{minipage}{12cm} %
    \small In these experiments, our goal is to forecast a sequence of values $Y=\{x_{t+1},... ,x_{t+12}\}$. We show the scores for $x_{t+3},\,x_{t+6},\text{ and } x_{t+12}$ for simplicity. %
  \end{minipage}%
}\\
\end{tabular}
\label{tab:multi step forecasting}
}
\vspace{-0.7cm}
\end{center}

\end{table}

\begin{table}[ht]
\renewcommand\arraystretch{1.05}         
\renewcommand\tabcolsep{2.0pt}
\caption{Experiments on spatial-temporal forecasting datasets (mean scores).}
\center
\scalebox{0.7}{
\begin{tabular}{c|ccc|ccc|ccc}
\toprule
\multicolumn{1}{c|}{Dataset} & \multicolumn{9}{c}{METR-LA}  \\ \hline

\multicolumn{1}{c}{} & \multicolumn{3}{|c|}{Horizon 1$\sim$3 } & \multicolumn{3}{|c|}{Horizon 1$\sim$6} & \multicolumn{3}{|c}{Horizon 1$\sim$12} \\

\hline
Methods/Metrics & MAE & RMSE & MAPE & MAE & RMSE & MAPE & MAE & RMSE & MAPE\\ \hline
StemGNN & 2.56& 5.06 & 6.46\% & 3.01 & 6.03 & 8.23\% & 3.43 & 7.23 & 9.85\% \\
STGNN & 2.62 &  4.99 &  6.55\%  & 2.98 &  5.88  & 7.77\% &  3.49  & 6.94 &  9.69\% \\
A2GNN & \textbf{2.43} & \textbf{4.44} & \textbf{6.12\%} & \textbf{2.65} & \textbf{5.05} & \textbf{6.93\%} & \textbf{2.92} & \textbf{5.83} & \textbf{8.05\%} \\ \hline
& \multicolumn{9}{c}{PEMS-BAY}\\ \hline
StemGNN& 1.23 & 2.48 & 2.63 & - & - & - & - & - & - \\
STGNN&  1.17 &  2.43 &  2.34\%  & 1.46 &  3.27 &  3.09\%  & 1.83 &  4.20 &  4.15\% \\
A2GNN& \textbf{1.08} & \textbf{2.14} & \textbf{2.23}\% & \textbf{1.29} & \textbf{2.75} & \textbf{2.81\%} & \textbf{1.52} & \textbf{3.42} & \textbf{3.42\%} \\ \bottomrule  
\multicolumn{10}{c}{%
  \begin{minipage}{12cm} %
    \small As STGNN and StemGNN published the mean score of all predicted horizons in their paper. Thus, we show the mean score (e.g. Horizon 1$\sim$3 represents the mean score of horizon $1,2,3$). %
  \end{minipage}%
}\\
\end{tabular}
\vspace{-2cm}

}
\label{tab:multi step forecasting (mean score)}
\end{table}

\subsubsection{Result Comparison on Spatial-temporal Dataset}
In order to further verify the performance of our model in the traditional spatial-temporal forecasting problem, we make further experiments on $2$ well known spatial-temporal datasets in traffic. 
We compare the performance of A2GNN with above-mentioned baseline methods, and these methods can not be employed to model time series problem without pre-defined graph.

We present the results on spatial-temporal forecasting tasks in Table \ref{tab:multi step forecasting} and compares corresponding performance of A2GNN with other spatial-temporal graph neural network methods. From this table, we can see that A2GNN achieves the best performance in terms of both RMSE and MAPE over almost all steps. In particular, A2GNN reduces the RMSE by $2.2\%$, $3.3\%$, $4.3\%$ on the METR-LA dataset and $3.0\%$, $4.4\%$ and $4.6\%$ on PEMS-BAY dataset when horizon is set to $3$, $6$ and $12$, respectively. 

Our method achieves the fastest training and inference time compared with other graph based models. The details are shown in appendix.

\subsection{Ablation Study}
\label{sec:ablation}
To gain a better understanding of the effectiveness of A2GNN's key components, we perform ablation studies through the time series forecasting task on Soloar-Energy dataset as well as the spatial-temporal forecasting task on META-LA dataset. The settings are summarized below:
\begin{itemize}
\item \textbf{A2GNN} is A2GNN method.
\item \textbf{w/o AGL} is A2GNN without auto graph learner.
\item \textbf{w/o $A^p$} is A2GNN without pre-defined relation.
\item \textbf{w/o ARL} represents A2GNN  attentional relation learner, which is replaced by $\rm{concatenate}$ operation.
\item \textbf{w/o A2} represents A2GNN without auto graph learner and attentional relation learner.
\end{itemize}

\begin{table}[t]
\caption{Ablation study.}
\renewcommand\arraystretch{1.05}         
\centering
\renewcommand\tabcolsep{3.0pt}
\scalebox{0.7}{
\begin{tabular}{c|cc|cc|cc|cc|cc}
    \toprule
& \multicolumn{4}{c}{Solar-Energy}  & \multicolumn{6}{|c}{METR-LA} \\
 \hline
 & \multicolumn{2}{c}{RSE$\downarrow$}  & \multicolumn{2}{|c|}{CORR$\uparrow$} & \multicolumn{2}{c}{MAE$\downarrow$}  & \multicolumn{2}{|c|}{RMSE$\downarrow$} &   \multicolumn{2}{c}{MAPE$\downarrow$} \\ \hline
Methods & Valid & Test & Valid & Test & Valid & Test & Valid & Test & Valid & Test\\ \hline
w/o AGL & 0.37 & 0.35     & 0.94 & 0.93 & 2.73 & 2.97     & 5.33 & 5.99     & 7.57\% & 8.27\%      \\
w/o $A^p$ & - & - & - & - & 2.75  & 2.98     & 5.29  & 5.97    & 7.53\%  & 8.17\%  \\
w/o ARL & 0.33 & 0.30     & 0.95 & 0.95 & 2.73 & 2.98     & 5.26 & 5.92     & 7.60\% & 8.28\%      \\ 
w/o A2  & 0.39 & 0.37     & 0.94 & 0.93 & 2.82 & 3.05     & 5.35 & 6.04     & 7.87\% & 8.57\%  \\ \hline
A2GNN & \textbf{0.32} & \textbf{0.28} & \textbf{0.95} & \textbf{0.95} & \textbf{2.71} & \textbf{2.93}     & \textbf{5.23} & \textbf{5.82}     & \textbf{7.50\%} & \textbf{8.14\%}      \\
 \bottomrule
\end{tabular}}

\label{tab:ablation study}

\end{table}


To perform the studies for time series forecasting tasks, we run experiments on the Solar-Energy dataset with $t_{out} = 12$. To perform the ablation study on spatial-temporal forecasting tasks, we report the experiments on the METR-TA dataset on all $12$ steps. In each experiment, the model is trained for $50$ epochs, and $10$ repeated runs ensure the reliability.
Table \ref{tab:ablation study} show the performance in terms of evaluation scores on both validation and test sets. From the table, we can see that, without auto graph learner, the performance drops drastically, which indicates that our auto graph learner plays an indispensable role in A2GNN for achieving more accurate forecasting. 
Similarly, the attentional relation learner is also responsible for a considerable performance gain by A2GNN. 
Specially, in spatial-temporal forecasting task, pre-defined graph/relation by human knowledge is a further information to be employed, which can also influence the model performance.
In the end, the experiment without auto graph learner and attentional relation learner prove the superiority of our overall approach.

\textit{The Effect of Neighbor Amount: }
Auto graph Learner (AGL) is designed to discover the implicit relation. In AGL, the $C$ is the key factor to control the neighbor amount for each node. Therefore, we set multiple $C$ to study the influence. As shown in Table \ref{tab:Effect of neighbor amount}, $C$ value that is too large or too small will make the model's performance worse.
A reasonable $C$ value will have a big improvement compared with the w/o AGL. Especially, the implicit relation discovered by AGL have a significant for time series forecasting task.

\begin{table}[]
\caption{Effect of neighbor amount. $C$ is the number of edges for each node in AGL.}
\renewcommand\arraystretch{1.05}         
\centering
\renewcommand\tabcolsep{3.0pt}
\scalebox{0.75}{
\begin{tabular}{c|cc|cc|cc|cc|cc}
    \toprule
& \multicolumn{4}{c}{Solar-Energy}  & \multicolumn{6}{|c}{METR-LA} \\
 \hline
 & \multicolumn{2}{c}{RSE$\downarrow$}  & \multicolumn{2}{|c}{CORR$\uparrow$} & \multicolumn{2}{|c}{MAE$\downarrow$}  & \multicolumn{2}{|c|}{RMSE$\downarrow$} &   \multicolumn{2}{c}{MAPE$\downarrow$}\\
 \hline
 $C$ & Valid & Test & Valid & Test & Valid & Test & Valid & Test & Valid & Test\\ \hline
 1 & 0.349  & 0.320  & 0.951  & 0.946  & 2.759  & 2.960  & 5.328  & 5.915  & 7.54\%  & 8.18\% \\
 3 & 0.337  & 0.302  & 0.954  & 0.952  & 2.725  & 2.945  & 5.233  & 5.848  & 7.50\%  & 8.16\% \\
 10 & 0.327  & 0.296  & 0.956  & 0.954 & \textbf{2.715} & \textbf{2.935}     & 5.233 & \textbf{5.825}     & \textbf{7.50\%} & \textbf{8.14\%} \\
 15 & \textbf{0.322}  & \textbf{0.289}  & \textbf{0.958}  & \textbf{0.957}  & 2.715  & 2.974  & 5.202  & 5.906  & 7.35\%  & 8.03\%\\
 30 & 0.333  & 0.302  & 0.954  & 0.952  & 2.720  & 2.953  & \textbf{5.210}  & 5.860  & 7.58\% & 8.22\%\\
\bottomrule
\end{tabular}}
\label{tab:Effect of neighbor amount}

\end{table}

\subsection{Interpretability Analysis}
\subsubsection{Analysis of Attentional Relation Learner}
To further reveal the effectiveness of the attentional relation learner, we provide intensive interpretability analysis on it (the pre-defined relation branch is disabled). The corresponding experiments are conducted on the META-LA dataset. 



\begin{figure}[]
    \centering
    \includegraphics[width=0.32\textwidth]{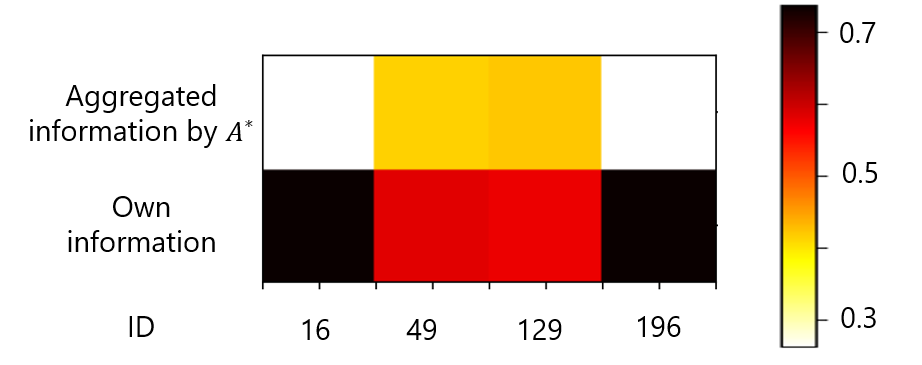}
    \caption{Visualization for the attention coefficients (over own information and neighbor information by AGL) learned by attentional relation learner for station 16, 49, 129, 196. }
    \label{fig:attentional relation learner attention matrix}
\end{figure}

\begin{figure}[]
    \centering
    \includegraphics[width=0.45\textwidth]{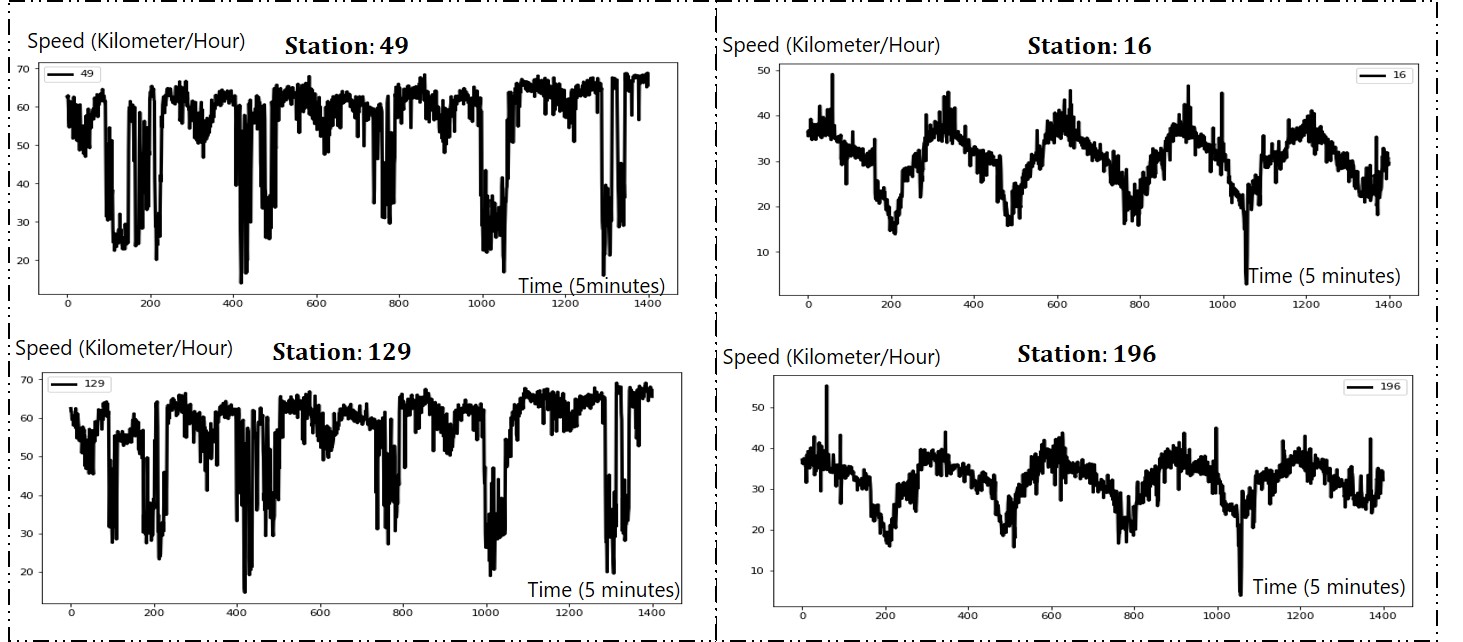}
    \caption{Station speed visualization for $49,129,16,196$. The speed curves of station $16$ and $196$ are similar to the sine waveform, while the speed curves of station $49$ and $129$ seems not stable.}
    \label{fig:attentional relation learner station curve}
\end{figure}

To show the relation attentions learned by attentional relation learner, we first sample some stations in the dataset, each of which corresponds to a node in the graph, and plot their attention coefficient on different relations in Figure \ref{fig:attentional relation learner attention matrix}. From this figure, we can see that various stations pay variant attention to different relations. For example, station $16$ and station $196$ pay more attention to identity adjacency matrix, i.e. they are more concerned about their own information. 
Furthermore, Figure \ref{fig:attentional relation learner station curve} visualizes stations with greater attention on the automatically learned adjacency matrix, i.e., station $49$ and $129$, on the left compared with stations with more concentrated attention on the identity matrix, i.e., station $16$ and $196$, on the right. It is obvious that the curves of station $49$ and $129$ are not stable with a couple of randomly occurred sudden drops. The curves of station $16$ and $196$, on the other hand, have a very clear and stable change pattern, where both the peak and the valley appear alternately and periodically. Therefore, it is much easier for station $16$ and $196$ to make accurate enough forecasting mainly based on their own information.

\subsubsection{Analysis of Auto Graph Learner}
\begin{figure}[t]
    \centering
    \includegraphics[width=0.4\textwidth]{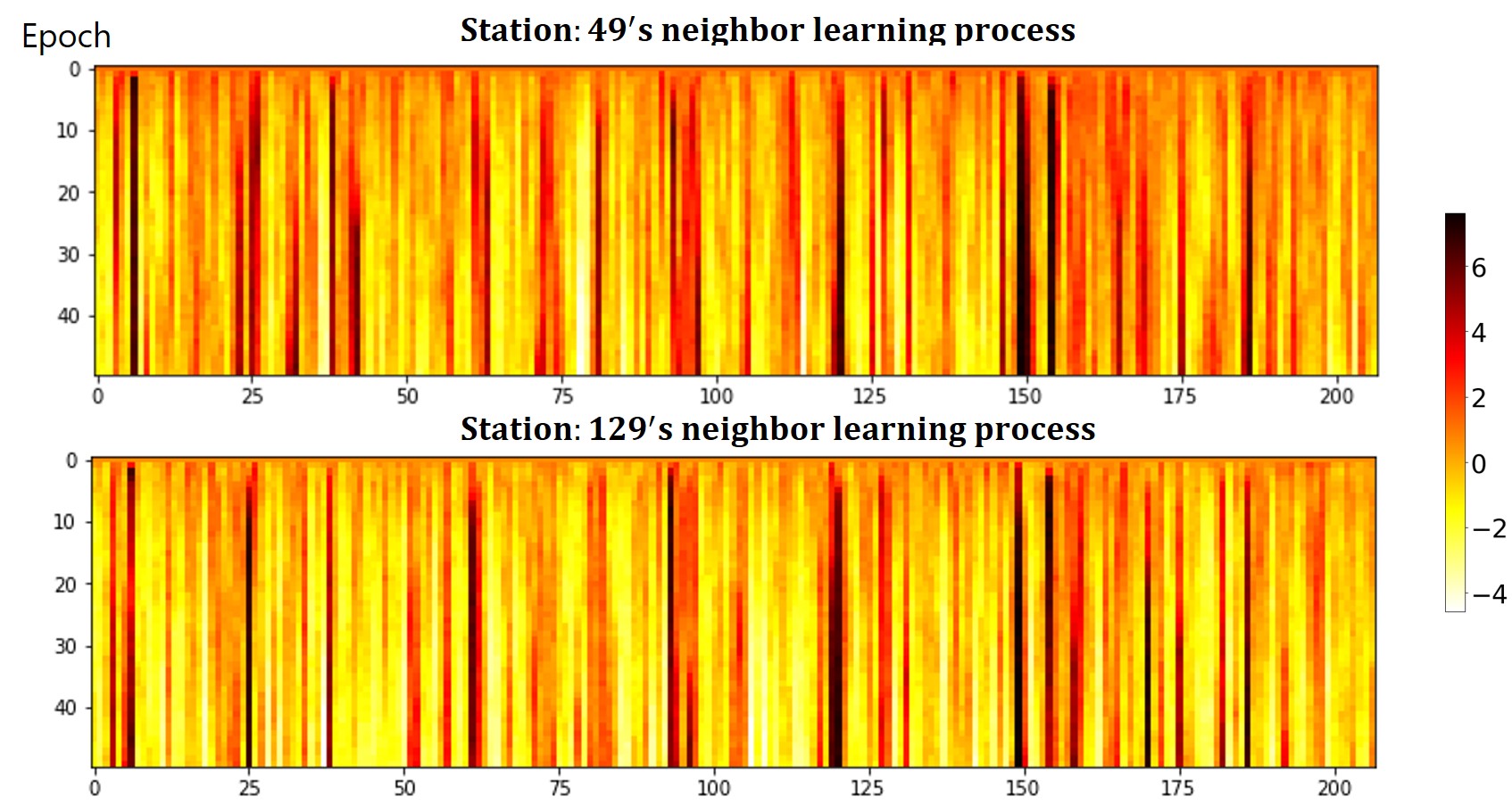}
    \caption{The learning process of auto graph learner for station $49,129$. }
    \label{fig:auto graph learner learning process}
\end{figure}

\begin{figure}[h]
    \centering
    \includegraphics[width=0.5\textwidth]{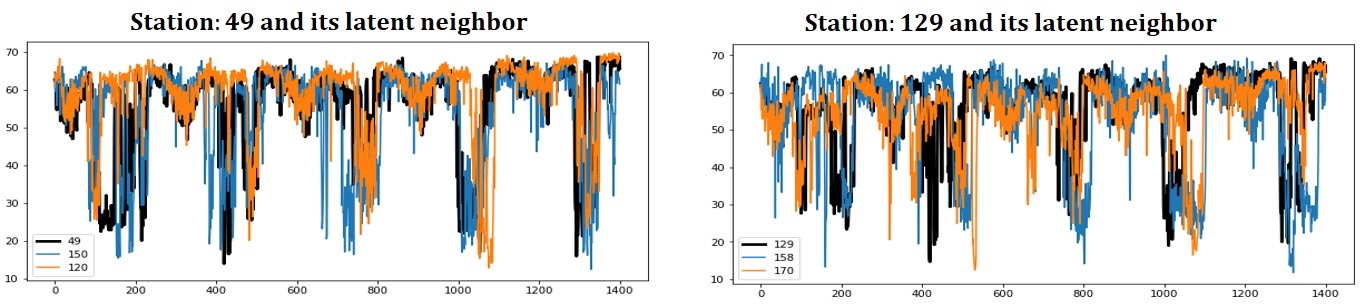}
    \caption{Speed visualization of station and its related neighbors learnt by auto graph learner. As we can see, the station and its implicit/learnt neighbors have a similar trend, which will improve the robustness. }
    \label{fig:top related node curve for station49}
\end{figure}

In this section, we make a further analysis on whether the learning process of auto graph learner is reasonable or not and whether the extra useful information exists in the learned neighbors from auto graph learner. In the first place, we visualize the correlation between station $49,129$ and all other stations during the training process in figure \ref{fig:auto graph learner learning process}. As we can see, as the training process goes on, the correlations between station $49$ and some stations, such as station $150,120$, become stronger and stronger (from light to dark in color). Moreover, we visualize station $49,129$'s neighbors with strong correlation learned from auto graph learner in figure \ref{fig:top related node curve for station49}. As we can see, the neighbor of station $49,129$ look like a similar curve of station $49,129$. The information aggregation of neighbors will enhance the predicted results (i.e. improve the robustness) of current station.
Thus, the learned neighbors can significantly improve the forecasting of station itself.

\section{Conclusions and Future Work}
In this paper, we propose attentional multi-graph neural network with automatic graph learning (A2GNN). Compared with previous studies, our framework can automatically learn sparse relation by using Gumbel-Softmax with facilitating each node to dynamically pay more attention to preferred relation graphs. Experiments on a couple of real-world datasets have demonstrated the effectiveness of A2GNN on a variety of time series forecasting tasks.


\small
\bibliography{aaai22}

\appendix
\section{Appendix}
\subsection{Dataset and Experimental Settings}

\begin{itemize}
\item Solar-Energy: The dataset from the National Renewable Energy Laboratory contains the solar power output.
\item Traffic: The dataset from California Department of Transportation contains road occupancy rates.
\item Electricity: The dataset from the UCI Machine Learning Repository contains electricity consumption.
\item METR-LA: The dataset from the Los Angeles Metropolitan Transportation Authority contains average traffic speed.
\item PEMS-BAY: The dataset from California Transportation Agencies (CalTrans) contains average traffic speed.
\end{itemize}

\begin{table}[htbp]

\caption{Dataset statistics. $A^p$ represents that the dataset has pre-defined relation or not. Samples is the sample number for every node. In and Out are the length of input and output respectively. }
\begin{center}
\scalebox{0.8}{
\begin{tabular}{c|cccccc}
\toprule
Datasets & $A^p$ & Nodes $\times$ Samples & Sample Rate & In & Out\\
\hline
Solar-Energy &  \ding{54} & 137 $\times$ 52,560 & 10 minutes & 168 & 1\\
Traffic & \ding{54} & 862 $\times$ 17,544  & 1 hour & 168 & 1\\
Electricity & \ding{54} & 321 $\times$ 26,304 & 1 hour & 168 & 1\\
\hline
METR-LA & \ding{52} & 207 $\times$ 34,272 & 5 minutes & 12 & 12\\
PEMS-BAY & \ding{52} & 325 $\times$ 52,116 & 5 minutes & 12 & 12\\
\bottomrule

\end{tabular}
\label{tab: dataset statistic}
}
\end{center}
\end{table}

For the training/validation/test data split, we follow the previous studies. Particularly, for the first $3$ datasets, we split them into training, validation and test set with ratio $6:2:2$ in chronological order according to \citep{mtgnn,11,graphwavenet}, while the remain $2$~datasets with ratio $7:1:2$ following \citep{mtgnn,gman}.

\textit{Evaluation Metrics: }To evaluate the model performances, we adopt five metrics, which are Mean Absolute Error (MAE), Root Mean Squared Error (RMSE), Mean Absolute Percentage Error (MAPE), Relative Squared Error (RSE) and Empirical Correlation Coefficient (CORR). The mathematical formulas for these metrics are listed as follows:


\begin{small}
\begin{eqnarray}
			&RSE(Y,\hat Y)&= \frac{\sum_{i=1}^n (y_i-\hat{y_i})^2}    {\sum_{i=1}^n (y_i-\overline{y})^2} ,\\
			&CORR(Y,\hat Y)&= \frac{\sum_{i=1}^n (y_i-\overline{y}) \times  (\hat{y_i}-\overline{\hat{y}}) }    {\sqrt{\sum_{i=1}^n (y_i-\overline{y}) ^2}    \sqrt{\sum_{i=1}^n (\hat{y_i}-\overline{\hat{y}}) ^2} }   ,\\
			&MAE(Y,\hat Y)&=\frac{1}{n}  \sum_{i=1}^n |y_i-\hat{y_i}|,\\
			&RMSE(Y,\hat Y)&=\sqrt{\frac{1}{n}  \sum_{i=1}^n (y_i-\hat{y_i})^2},\\
			&MAPE(Y,\hat Y)&=\frac{100\%}{n}  \sum_{i=1}^n   \left|  \frac{y_i-\hat{y_i}}  {y_i} \right|,
\end{eqnarray}
\end{small}
where $n$ is the number of evaluation samples, $\hat{Y}$ is the set of prediction results and $Y$ is the set for ground truth. Note that the evaluation is for the forecasting of timestamp $t+t_j$ based on the $t$ status, $t$ is the current timestamp and $t_j\in\{1,2,3,...,t_{out}\}$.

\textit{Experimental Setup in Training: } We train the model with Adam \citep{adam} optimizer with gradient clip value $5$. The learning rate for the parameters in the auto graph learner part is set to $0.01$. For parameters in other components they are set to $0.001$. Dropout \citep{dropout} in graph neural network is set to $0.3$. Temperature $\tau$ for Gumbel-Softmax operation in auto graph learner is set to $0.5$. Other dataset-specific hyper-parameters of A2GNN are summarized in Table \ref{tab: model configuration}. LSTM$_{out}$ represents the output dimension of LSTM. $L$ represents the layer number of graph neural network. GNN$_{out}$ means the output dimension of graph neural network. In attentional relation learner, the hidden dimension of linear for query, key is both $128$, and $d_{value}$ means the hidden dimension $W_{value}$ in attentional relation learner. $C$ is the sample number of Gumbel-Softmax operations, which represents the number of neighbors.

\begin{table}[htbp]

\caption{Model Configuration.}
\begin{center}
\scalebox{0.75}{
\begin{tabular}{c|cccccc}
\toprule
Dataset &  LSTM$_{out}$ & $L$ & GNN$_{out}$ & $d_{value}$& $C$\\
\hline
Solar-energy &  16 & 2& 32 &128& 15\\
Traffic  & 16 &2&  128 & 256 &15 \\
Electricity& 16 & 2& 32 & 128 &15 \\
\hline
METR-LA & 16 & 2& 128 & 256 &15\\
PEMS-BAY & 32 & 2& 256 & 512 &15\\
\bottomrule

\end{tabular}
\label{tab: model configuration}
}
\end{center}
\end{table}

\subsection{Baseline Methods for Comparision}
As we mentioned above, the biggest difference between time series forecasting task and spatial-temporal forecasting task lies in whether there exists a pre-defined relation. We make a detailed introduction about the baseline methods:

\parag{Time Series Forecasting}
\begin{itemize}
\item \textbf{AR}: An auto-regressive model.
\item \textbf{VAR-MLP \citep{7}}: A hybrid model of multilayer perception (MLP) and auto-regressive model (VAR) .
\item \textbf{GP \citep{8}\citep{9}}: A gaussian process time series model, which is a Bayesian nonparametric generalisation of discrete-time nonlinear state-space model.
\item \textbf{RNN-GRU}: A recurrent neural network with fully connected GRU hidden units.
\item \textbf{LSTNet \citep{10}}: A model which combines convolutional neural network and recurrent neural network.
\item \textbf{TPA-LSTM \citep{11}}: An attention-based recurrent neural network, which learns to select the relevant time series.
\item \textbf{MTGNN \citep{mtgnn}}: A uni-directional graph neural net, which employs temporal convolution and dilated inception layer.
\end{itemize}

\parag{Spatial-temporal Forecasting}
\begin{itemize}
\item \textbf{DCRNN \citep{dcrnn}}: A diffusion convolutional recurrent neural network, which combines diffusion graph convolutions with recurrent neural networks.
\item \textbf{STGCN \citep{stgcn}}: A spatial-temporal graph convolutional network, which incorporates graph convolutions with 1D convolutions.
\item \textbf{Graph WaveNet \citep{graphwavenet}}: A spatial-temporal graph convolutional network, which integrates diffusion graph convolutions with 1D dilated convolutions.
\item \textbf{ST-MetaNet \citep{stmetanet}}: A sequence-to-sequence network, which employs the meta knowledge to generate edge attentions.
\item \textbf{GMAN \citep{gman}}: A graph multi-attention network with spatial and temporal attentions.
\item \textbf{MRA-BGCN \citep{mrabgcn}}: A multi-range attentive bicomponent GCN.
\item \textbf{MTGNN \citep{mtgnn}}: A uni-directional graph neural net, which employs temporal convolution and dilated inception layer.
\item \textbf{STGNN \citep{STGNN}}: A graph neural network with learnable positional attention.
\item \textbf{StemGNN \citep{StemGNN}}: A spectral temporal graph neural nework, which combines both Graph Fourier Transform (GFT) and Discrete Fourier Transform (DFT).  
\end{itemize}

\subsection{Computation Time}
To test the computational efficiency our our model, we compared the training time with other graph based models on the METR-LA dataset in Table \ref{tab:computation time}. STGCN incorporates graph convolutions with 1D convolutions, therefore, the training time is fast. A2GNN have a same training time as STGCN. For inference, DCRNN and STGCN employ the recursive decoder to predict, and the inference time is too slow. Other models like Graph WaveNet, MTGNN, STGNN, StemGNN and our method generates all predictions in on run, which is more engineering friendly.

\begin{table}[!ht]
\caption{The computation cost on the METR-LA dataset.}
\renewcommand\arraystretch{1.05}         
\renewcommand\tabcolsep{3.0pt}

\begin{center}
\scalebox{0.6}{
\begin{tabular}{cccccccc}

\toprule

Model           & DCRNN     & STGCN & Graph Wave &MTGNN & STGNN & StemGNN & A2GNN \\
Train(s/epoch)  & 249.31    & \textbf{19.10} &53.68 &48.40   &56.37 &94.77 &24.5 \\
Inference (s)   & 18.73     & 11.37& 2.27&          1.89&  2.38 & 4.25&  \textbf{0.98}\\
\bottomrule
\end{tabular}}
\end{center}
\label{tab:computation time}

\end{table}

\end{document}